# Obstacle Avoidance Using Stereo Camera

Akkas Uddin Haque, Ashkan Nejadpak
Department of Mechanical Engineering
University of North Dakota

*Abstract*—**In this paper we present a novel method for obstacle avoidance using the stereo camera. The conventional obstacle avoidance methods and their limitations are discussed. A new algorithm is developed for the real-time obstacle avoidance which responds faster to unexpected obstacles. In this approach the depth map is divided into optimized number of regions and the minimum depth at each section is assigned as the depth of that region. A fuzzy controller is designed to create the drive commands for the robot/quadcopter. The system was tested on multiple paths with different obstacles and the results demonstrated the high accuracy of the developed system.**

## I. Introduction

Why obstacle avoidance? To prevent robot/quadcopter from crash and failure. In many navigation systems, the obstacle avoidance problem is solved using Ultrasonic sensors. The ultrasonic measurements suffer from the following limitations; they are not convenient for mapping or other tasks that require high accuracy. They are very sensitive to noise and therefore not recommended for robots operating in narrow paths and will fail detecting the unexpected obstacles [1]. As humans use vision for obstacle avoidance for navigating, here the same vision based algorithm is developed for obstacle avoidance which uses the stereo camera to create the depth maps.

There are a lot of quadcopters or small form-factor UAV's used by hobbyists and professionals alike today. Even though flying UAV's outdoors is trivial and requires little to no experience, owing to the fact that sensors like GPS do not work indoors, flying them indoors requires more skill and dexterity to avoid crashing into walls and other obstacles. Current obstacle avoidance algorithms based on stereo systems are either too computationally intensive or are able to make just basic decisions like going left, right and forward. We propose an active control method using fuzzy logic to detect obstacles that would be both fast and would accurately avoid obstacles. It would also be able to deal with more control parameters like top-left, top-right, bottom-left and bottom-right.

Since the depth map has a lot of information that would take a significant amount of time to process, we propose to divide the map to multiple regions. Each of these regions would contain the value of the closest (highest intensity value) object in the region. Reducing the disparity map to a grid of regions containing the closest value of depth would increase response time while also retaining enough information to avoid obstacles.

## II. Background/Literature Review

Obstacle Avoidance using stereo vision with the depth processing done on a FPGA has been explored in [2]. Captured stereo images are rectified using sum of absolute differences to produce the disparity image. The disparity image is then divided into 3 regions, left, middle and right and decisions are made based on the obstacle detection on the three regions. The FPGA implementation enables fast computation of the disparity map.

While the above implementation worked with rectangular regions aligned horizontally, the disparity map is divided into 3 regions aligned vertically in [3]. One interesting feature of this paper is that it proposes metric to measure the reliability of each point in the disparity image. It does this by giving a lower score to texture-less regions and higher score to more textured regions.

Both the previous systems were implemented on ground based systems and were thus able to reliably perform with the relatively simple system. These systems would not work for a complex system like quadcopters. Olivares-Mendez et. al propose a different approach to obstacle avoidance. They present a visual fuzzy servoing system for detecting and avoiding obstacles [4]. Their system however, does not estimate the distance to the obstacle and the obstacle is infact a known object.

In contrast to the previous systems, our system defines 9 regions that are optimized to suit the quadcopter dynamics. Moreover, a fuzzy controller that has as inputs the closest distance to objects viewed in each region, has been implemented to produce a smoother response the control inputs.

## III. System Setup

For the efficient and fast computation of depth from stereo images, on-board a small form-factor computer, a system had to be chosen that is capable of massively parallelizing the computation of disparity/depth. To this end, the Jetson TX1 from NVIDIA was selected, which has a 256 core GPU capable of 1TFLOPS theoretical performance and a compute capability of 5.2. A stereo camera from Zed Camera systems is used which has a stereo baseline of 120mm. The lens has a wide-angle field of view of 110 degrees.

To be able to leverage the support of the GPU, CUDA was installed on the Jetson TX1. CUDA is a parallel computing platform and application programming interface (API) model created by Nvidia.[5] It allows software developers and software engineers to use a CUDA-enabled graphics processing unit (GPU) for general purpose processing – an approach termed GPGPU (General-Purpose computing on Graphics Processing

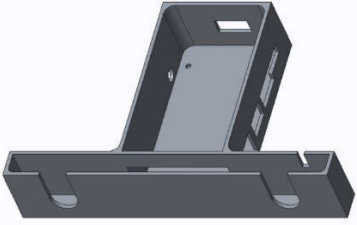

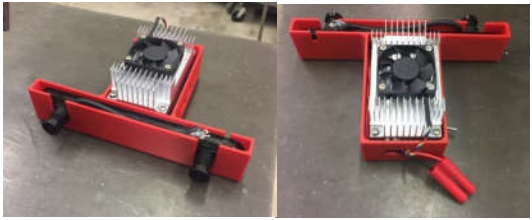

Fig. 1. The container designed using CREO

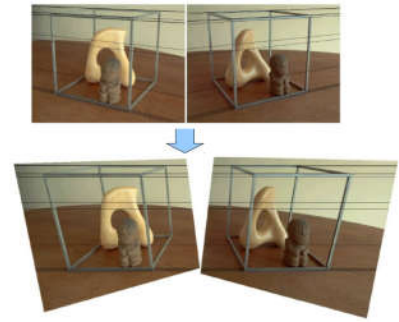

Fig. 2. Top: Calibrated Images,
Bottom: Stereo Rectified Images

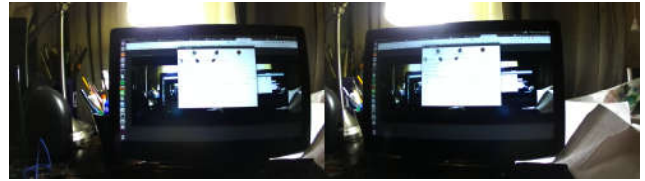

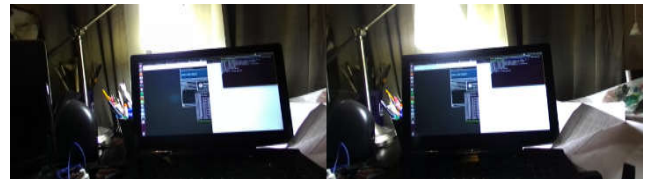

Fig. 3. Top: Raw images from ZED Camera,
Bottom: Stereo Rectified Images

Units). The CUDA platform is a software layer that gives direct access to the GPU's virtual instruction set and parallel computational elements, for the execution of compute kernels.

Finally, a setup consisting of the stereo camera, coupled with the Jetson and a carrier board, Orbitty from Connect Tech Inc. was built to ensure modularity and easy mounting of the system to a quadcopter. Care was taken to reduce the amount of material in the construction to ensure the structure would be both lightweight, to enable the minimize the load on the quadcopter, and robust enough to withstand the impact of a crash during flight.

## IV. CONTAINER DESIGN

In order to collect data, and for future research which is an autonomous flight, a container case is designed which holds all the systems components. The corresponding dimensions and locations for all inputs, camera, switch and power inlet are measured. The container shown in figure below, is designed using PTC Creo software.

To make sure the container will not fail due to the corresponding weight of the costly system's components, a stress analysis is performed on it using Ansys software. The results demonstrated an acceptable design. The container was examined under 20N force acting on the lower plane of the container which is ten times more than the actual corresponding force. The material propertise were modified for the Acrylonitrile Butadiene Styrene (ABS) material; this material is less brittle and more ductile compared to other 3d printing mateials such as PLA filament and etc. The maximum deformation was found to be less that a millimeter (0.7 mm), and the maximum stress is less than 5% of the maximum yield strength. The final 3d printed product is shown on the figure below.

Moreover, modal analysis was performed on the empty case to avoid resonance frequencies (First three modes are 155, 193, 210 Hz), specially if it is attached to a quadcopter for future use.

## V. DEPTH COMPUTATION

Depth is computed using a calibrated stereo camera. The ZED camera produces synchronized right and left images as a single image. The raw images received from the cameras are uncalibrated. To be able to produce depth images, the cameras must be calibrated in two stages: single camera calibration and stereo camera calibration.

Each of the cameras in the stereo pair were first calibrated to retrieve the intrinsic parameters of the cameras. The calibration was done based on the pinhole camera model as outlined in [6].

The second stage is to do stereo calibration. This stage basically involves finding the transformation between the cameras, making them a stereo pair. It also involves finding the fundamental matrix F, which encodes the transformation between the pixel coordinates in the image from one camera to the pixel coordinates in the second image. Once the fundamental matrix is known, the corresponding points in the two images can be related by

$$m_2^T F m_1 = 0 \qquad (1)$$

where $m_1$, $m_2$ are the corresponding pixel coordinates in the first and second images respectively. If one of the coordinates, $m_1$ in known and the corresponding point $m_2$ is unknown, the equation (1) defines a line, also called the epipolar line.

Unfortunately, using the fundamental matrix to scan for corresponding points in the second image is not very efficient computationally. This computation time is greatly reduced if the search is along the row/column. This is where stereo rectification comes in. Stereo rectification transforms the images so that the epipolar lines are lined up with the direction of image rows/columns also called scanlines. For a left-right camera pair, the scanlines are usually horizontal as shown in figure 2.

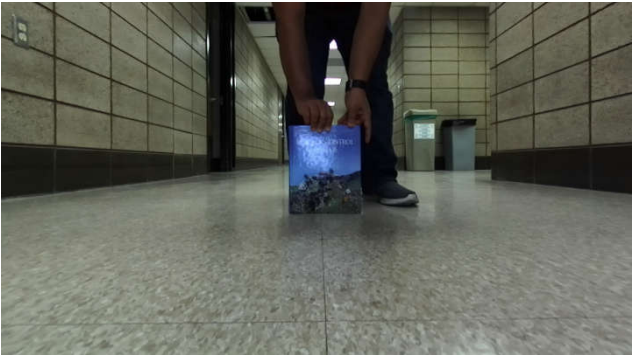

a)

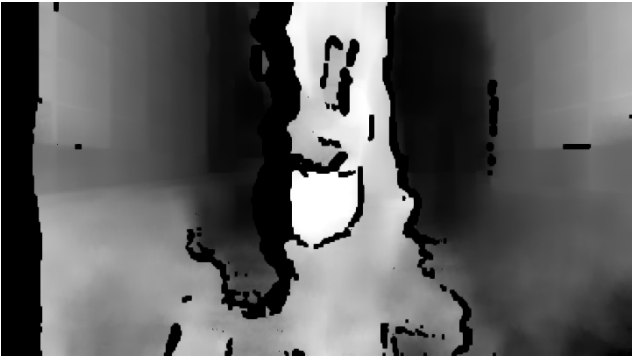

b)

Fig. 4. Refining of depth. Stereo images are taken at multiple locations in front of the camera and the depth at the center of the book is computed for each image. These are then used to create a lookup table a) Image of the book at one of the positions. b) The corresponding depth map

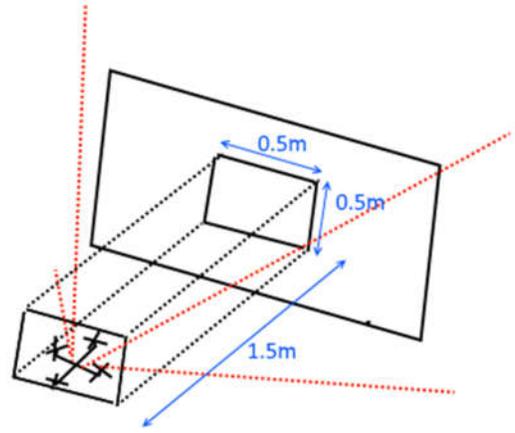

Fig. 5. Optimization of region size. A square region of side 0.5m at is distance of 1.5m from the camera center is selected as the size of the center region. This 0.5m square corresponds to 150 pixels' square in the image. The other regions are defined according to the ranges of the center region.

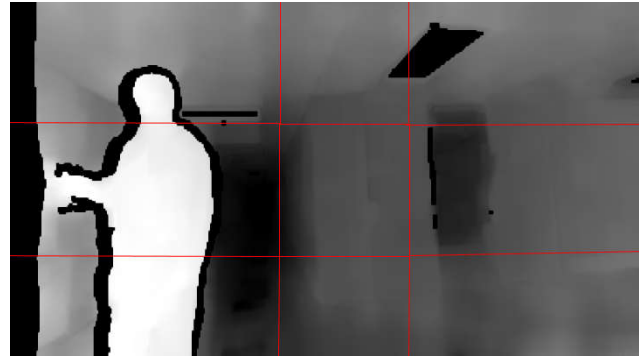

Disparity in stereo rectified images can then be easily computed by $x - x'$, where $x, x'$ are the distance between points in the image plane corresponding to the scene point in 3D and their respective camera centers. The relation between disparity and depth is given by

$$disparity = x - x' = \frac{B * f}{z}, \qquad (2)$$

where B is the distance between the two cameras, also called the baseline and $f$ is the focal length of the camera and $z$ is the distance of the point from the camera plane to the point examined. In short, above equation says that the depth of a point in a scene is inversely proportional to the difference in distance of corresponding image points and their camera centers. With this information, the depth of all pixels in the image pair can be calculated.

Since the computation of the disparity and depth for each pixel is not dependent on the other pixels in the image, the computation is massively parellelizeable. A CUDA kernel was written to this end that computes the disparity map from the set of stereo images by block matching. The use of the GPU for the computation of the depth map onboard the Jetson, resulted in an average performance improvement of 35 times than if done on the CPU.

VI. REFINING OF DEPTH

The depth calculated in the previous section does not represent the true depth observed in the real world. This was further refined by performing a manual calibration of the depth values computed and the true depth sampled at different distances from the camera. A series of images were recorded using the camera and an application was written to output the depth at any point clicked on the depth map produced from stereo pairs of images. The actual distance to a known object in the image was then correlated with the depth produced at the position of the object and a lookup table correlating the two distances was constructed. This was then used to find the true depth for the given computed depth.

## VII. DISTILLING THE DATA

The depth map produced has a lot of information that would be computationally intensive to process. Thus we propose a method to distill the data, retaining only the most relevant information, while still being able to perform obstacle avoidance maneuvers reliably. We propose dividing the depth map to 9 different regions namely, up, down, left, right, center, up-left, up-right, down-left, down-right as shown in fig???

The size of the center region was defined as the projection of a window of safe distance around the quadcopter onto a plane 1.5m in front of the image plane. This was chosen to be a .5x.5 meter square around the center of the quadcopter. This was computed experimentally by pointing the camera at a plane 1.5m away from the camera plane and recording the size in pixels of the .5x.5 meter box, drawn on this plane, makes on the image. This was found to be about 150 pixels wide by 150 pixels high. Thus, the center region is defined to be 150x150 pixels, at the center of the left image, since that is the image that we are calculating the disparity of the right image against.

The closest distance in each of these 9 regions is computed during the depth map creation itself. This required the addition of code in the CUDA kernel to find minimum depth in each region. Performing this computation within the kernel also enabled the computation of minimum distance within each region in real-time. The depths produced from this operation were then refined using the Look-Up Table created earlier, which are then passed onto the Fuzzy controller for further processing

## VIII. FUZZY CONTROLLER

In the previous section the depth map is divided into 9 regions, and a depth value is assigned for each region. Now, it is necessary to define a controller which creates the most efficient drive path. As discussed below, for this problem, a fuzzy controller will perform best.

First, let's compare the fuzzy sets versus the crisp sets. in a crisp set, an element is either a member of a set or not. Fuzzy sets on the other hand, allow elements to be partially in a set. Each element is given a degree of membership in a set. This membership value can range from 0 (not an element of the set) to 1 (a member of the set). It is possible to make some comparisons between fuzzy sets and statistical classifications and neural networks. Regarding the former, we argue fuzzy works with degree of membership, in other words similarity of an element to a class. Statistical classifications work with probability of being in a set. Probability involves crisp set theory and does not allow for an element to be a partial member in a class. The difference between fuzzy classifiers and neural network is that neural network is initialized in a random state while the membership functions of a fuzzy classifier can be initialized in a state close to the correct solution. Therefore, optimization of a fuzzy classifier is much faster than a neural network. However, the problem with fuzzy system is it is difficult to deal with too many features, membership functions and rules. Neural networks are more suited for large amounts of features and classes. The basic claim of fuzzy logic is that everything is a matter of degree. This is based on analysis of real world problems and the resulting paradox of applying conventional mathematics rooted in set theory. One of the most famous examples for comparing fuzzy logic with conventional mathematical analysis is the liar paradox [7]. This paradox states: Does the liar from Crete lie when he says all Cretans are liars. If he lies, he is actually telling the truth. If he is telling the truth, due to statement he made, he is lying. the reason conventional mathematical analysis fails here is the self-reference issue trending in these problems. While the fuzzy logic would obtain the result that the response lies between the lie and the truth.

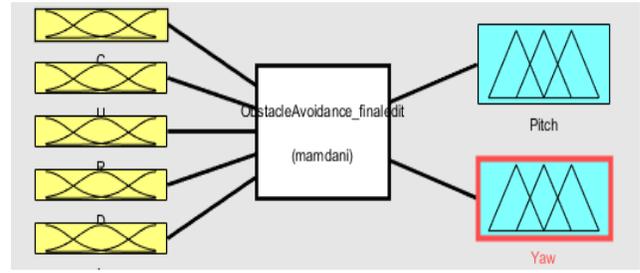

Fig. 6. The Mamdani fuzzy controller developed

Overall there are six steps that need to be completed to design a fuzzy inference system [8]. These steps are:

1. Determining the fuzzy rules,

2. Fuzzifying the inputs using input membership functions,

3. Combining the fuzzified inputs according to the fuzzy rules to establish a rule strength,

4. Finding the consequence of the rule by combining the rule strength and the output membership function,

5. Combining the consequences to get an output distribution,

6. Defuzzifying the output distribution.

There are two types of Fuzzy inference systems, Mamdani and Sugeno. For this problem, the mamdani method is used as it is intuitive, has widespread acceptance and is well suited to human input. In making a fuzzy rule, we use the concept of "and", "or" and "not". Therefore, we need to define the fuzzy combinations known as T-norms. The fuzzy "and" is written as:

$$u_{A \cap B} = T(u_A(x), u_B(x)), \qquad (3)$$

where, $u_A$ is the membership in class A and $u_B$ is membership in class B.

There are many ways to compute "and". The most common ones are Zadeh which takes the minimum of the two membership values and is the most common definition of the fuzzy "and". The second one is calculating the product of the two membership values. Similarly, the and function is defined as maximum of two membership values and calculating the difference between the sum of the two and the product of the membership values.

The results are calculated using two steps. First, the rule strength is calculated by combining the fuzzified inputs. The second part is clipping the output membership functions at the rule strength. More details about this process is discussed in our example in next section.

The last step is defuzzification of the outputs. There are two techniques available for defuzzification. First one is calculating the center of mass using the following equation:

$$z = \frac{\sum_{j=1}^{q} Z_j\, u_c(Z_j)}{\sum_{j=1}^{q} u_c(Z_j)}, \quad (4)$$

where, $Z$ is the center of mass, $u_c$ is the membership in class c at value $z_j$.

The second method is mean of maximum, given by

$$z = \sum_{j=1}^{l} \frac{z_j}{l}, \quad (5)$$

where z is the mean of maximum, $z_j$ is the point at which the membership function is maximum, l is the number of times the output distribution reaches the maximum level.

In this problem, the fuzzy controller is designed using MATLAB. First, the overall inputs and outputs are defined. It is clear that the inputs are the depths of the nine regions and the outputs are the path the robot would select. The main rule the robot should follow is that it has to choose the closest flight distances. This means that a code has to be developed that would first only consider the center region, and if it has to change its path, like when an obstacle is close to the center region, it would choose the main four directions (Right, left, Up, and Down) as they are closer and will not consider the extensive corners of the image yet. Then in the analysis section of the code developed, another fuzzy controller is considered which follows the same rules but it's basically 45 degrees rotated version of the first fuzzy controller to take the corners into consideration if the closest regions are to be avoided.

There are two outputs for this problem. The two outputs are amount of movement to the left or right (Yaw) or up and down (Pitch). Here the speed of the robot is considered to be constant and it's not considered as an output. For future research, the speed of the quadcopter could be considered as another output which will result in having the best optimized system. The input members are defines based on the corresponding distances. It is considered that the distance less than 0.75m are near, and the distances more than 2.25 are far, therefore, a scale of 1/3 is applied to the input members and the result is shown on the figure below. The outputs are defined with the same approach. Next, the fuzzy rules are combined and the rules are defined based on human language logics and the output results are modified based on these rules. For example, we define that if center is far and whatever distance the rest of regions are, go straight and the final values for pitch and yaw would therefore be zero. Overall there are seven rules defined for the fuzzy controller.

1. If center region is far, then the output values for pitch and yaw would be zero.
2. If center is near, and upper region is near, and the lower region is far, the output direction of the pitch is downward.
3. If center is near, lower region is near, and the upper region is far the output direction of the pitch is upward.
4. If center is near, right2 region is near and the left region is far, the direction of the output is toward left.

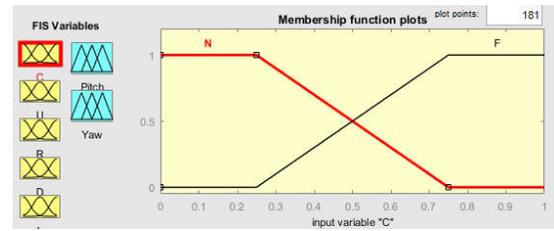

Fig.7 . The optimized input members designed.

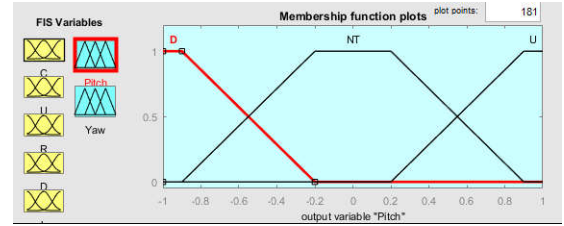

Fig 8. .Output members of the fuzzy inference

5. If center is near, left region is near, and the right region is far, the direction of the output is toward right.
6. If center is near and right region is near, the output direction is toward right.
7. If center is near and upper region is near, the output direction is upward.

The last step is defuzzifying the output distributions. Some examples of the final results are shown in the figures next page, the final values of pitch and yaw after defuzzification process are shown on the images. The figures on the next page display the final crisp output value after the defuzzification process. Figure 9 (a) demonstrates that when the center region is far, the controller does not take the other regions into consideration and would define the output command as move straight forward with no turn needed. For the rest of figures where the center region is near, the controller would calculate the final results based on the other input values.

In figure (b), the center region is near, based on the other region's depth values, the pitch would be upward as the upper region is further than the lower region and the yaw would be toward the left. In figure (c) the center, up, and down regions are all completely near. In this case the right and left regions are both far. Therefore, a rule is defined that in situations such as this one the robot would choose one of the two available paths (in this case the right path). The problem is, if this rule was not assigned, the center of mass would have been calculated close to zero and the controller would fail. Therefore, the controller was modified to avoid this type of failures.

For figures (d) the center value is between 0.25 and 0.75, and it is not in the area of absolute near or far distance. Therefore,

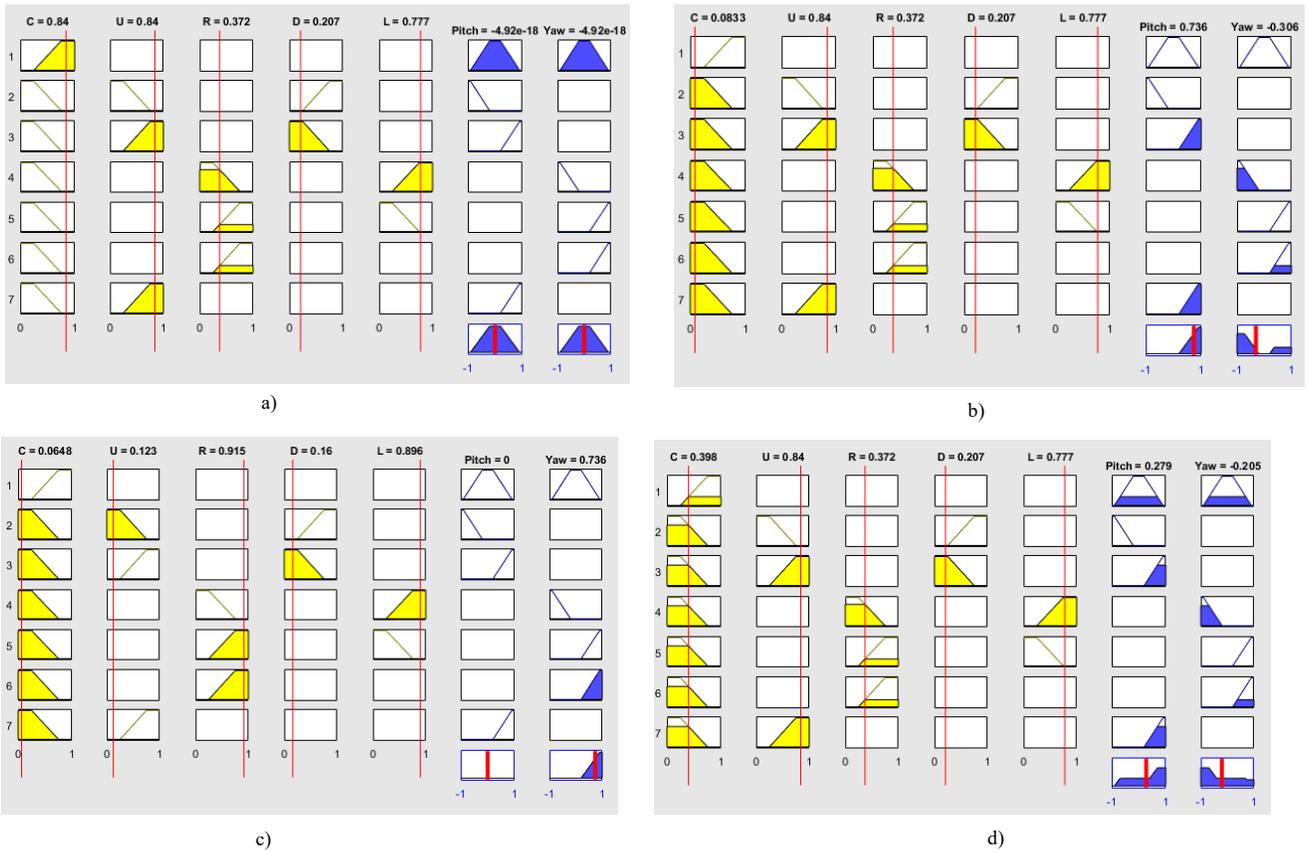

Fig 9. Fuzzy Rule Base

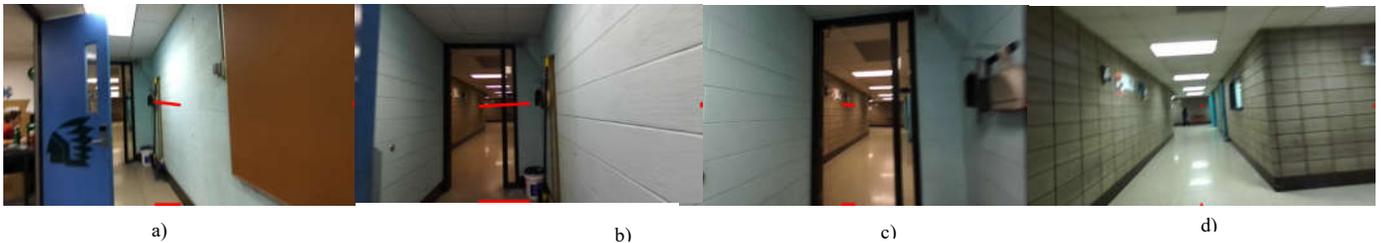

Fig 10. The controller navigating a particularly difficult and narrow passage. In a) the indicator points slightly towards the left, indicating that there is a space available a little to the left b) After crossing the door, the indicator points more towards the left as we have been veering towards the right. As the doorway comes into view (c) the pointer gets back to the center indicating that the path is clear ahead

the fuzzy controller has to consider the other regions as well. Comparing the values of Up and down regions, it is noticed that the upper region stands in far area while the down region is near. Same analysis is performed for left and right regions where the left region is further than the right one. As mentioned in the previous sections, for final output value the controller calculates the centroid of the sum of the maximum fuzzy outputs.

## IX. RESULTS

The fuzzy controller based obstacle avoidance mechanism implemented in could perform reliably in a variety of different settings. One such setting is shown in fig 10. The controller tries to navigate a narrow passage and succeeds in providing the right outputs. This is seen from the length and direction of the indicator. The length of the indicator indicates the amount of yaw and pitch necessary to move the quadcopter away from obstacles.

Fig 11 shows a case where the system fails. An obstacle that stays in the right region goes undetected until it is very close to the quadcopter. This sudden appearance of an object in the center region causes the controller to indicate a turn sharply to the top right corner to avoid the obstacle. In a real quadcopter test, this would have caused the quadcopter to crash.

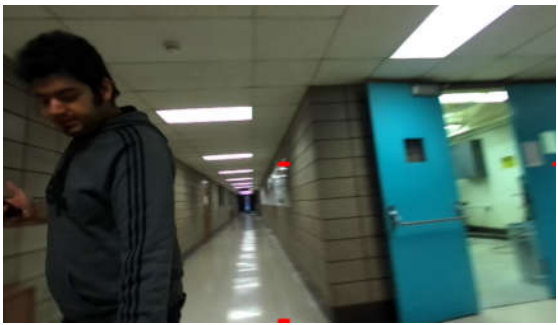 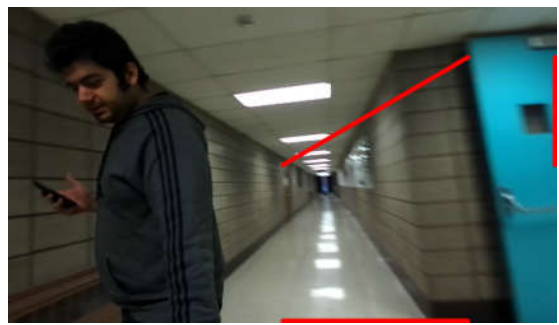

a)  b)

Fig 11. A case of failure: In (a) the obstacle is not detected in the center region so the controller provides no output. In (b) the obstacle has moved to the center region. Seeing that the object is so close in the center region, the controller indicates a sharp turn to the top right to move away from the object. In a real test, this would have caused the quadcopter to crash.

## X. Conclusions and Future Work

A fuzzy based obstacle avoidance strategy based on depths calculated using a stereo camera was formulated and tested using a live system. Future work would include incorporating a motion model into the fuzzy controller and also use the velocity of the quadcopter and previous outputs of the fuzzy controller as inputs, basically creating a feedback system.

The obstacle avoidance algorithm could be run alongside a SLAM algorithm to execute global localization and mapping with avoiding obstacles along the way.

## Acknowledgements

We thank Professor Stanlake for his invaluable assistance in helping us 3D print out model.